\documentclass{article}
\usepackage{graphicx} 
\usepackage{subcaption}
\usepackage{amsmath,amsthm}
\usepackage{hyperref,cleveref}
\usepackage{xcolor}
\usepackage{url}
\usepackage{todonotes}
\usepackage{enumitem}
\usepackage{authblk}
\usepackage{multirow}
\hypersetup{
    colorlinks,
    citecolor=black,
    filecolor=black,
    linkcolor=black,
    urlcolor=black
}
\newcommand{\brio}{BRIO}

\title{Evaluating AI fairness in credit scoring\\with the BRIO tool}
\date{}

\author[1,6]{Greta Coraglia$^*$}
\author[1,6]{Francesco A. Genco\thanks{These authors were supported by the Italian Ministry of University and Research through the project  PRIN n. 2020SSKZ7R BRIO ``Bias, Risk and Opacity in AI'', and the Project ``Departments of Excellence 2023-2027'' awarded to the Department of Philosophy ``Piero Martinetti'' of the University of Milan.}}
\author[2,4]{Pellegrino Piantadosi}
\author[5]{Enrico Bagli}
\author[5]{Pietro Giuffrida}
\author[3,6]{Davide Posillipo}
\author[1,6]{Giuseppe Primiero$^*$}

\affil[1]{Logic, Uncertainty, Computation and Information Lab, Department of Philosophy, University of Milan, Italy}
\affil[2]{Dipartimento di Fisica e Astronomia, University of Bologna, 
	Italy}
\affil[3]{AI Evolution Hub, Alkemy S.p.A., Milan, Italy}
\affil[4]{INFN, Sezione di Bologna,
	Italy}
\affil[5]{CRIF S.p.A., 
	Bologna, Italy}
\affil[6]{MIRAI S.r.l., Milan, Italy}

\begin{document}

\maketitle

\begin{abstract}

We present a method for quantitative, in-depth analyses of fairness issues in AI systems with an application to \emph{credit scoring}. To this aim we use \brio{}, a tool for the evaluation of AI systems with respect to social unfairness and, more in general, ethically undesirable behaviours. It features a \emph{model-agnostic bias detection module}, presented in \cite{DBLP:conf/beware/CoragliaDGGPPQ23}, to which a full-fledged unfairness risk evaluation module is added. As a case study, we focus on the context of credit scoring, analysing the UCI German Credit Dataset \cite{misc_statlog_(german_credit_data)_144}.
We apply the \brio{} fairness metrics to several, socially sensitive attributes featured in the German Credit Dataset, quantifying fairness across various demographic segments, with the aim of identifying  potential sources of bias and discrimination in a credit scoring model. We conclude by combining our results with a revenue analysis.

\end{abstract}

\section{Introduction}\label{sec:intro}


In recent years, the integration of Artificial Intelligence (AI) into various domains has brought forth transformative changes, especially in areas involving decision-making processes. One such domain where AI holds significant promise and scrutiny is credit scoring. 

Traditionally, credit scoring algorithms have been pivotal in determining individuals' creditworthiness, thereby influencing access to financial services, housing, and employment opportunities. 
The adoption of AI in credit scoring offers the potential for enhanced accuracy and efficiency, leveraging vast datasets and complex predictive models \cite{Guidolin2021}. Nevertheless, the inherently opaque nature of AI algorithms poses challenges in ensuring fairness, particularly concerning biases that may perpetuate or exacerbate societal inequalities. Fairness in credit scoring has become a paramount concern in the financial industry.
According to the the AI act and to the European Banking Authority guidelines---which state that ``the model must ensure the protection of groups against (direct or indirect) discrimination'' \cite{EBA2020}---ensuring fairness and the prevention/detection of bias is becoming imperative. Fairness is fundamental to maintaining trust in credit scoring systems and upholding principles of social justice and equality. Biases in credit scoring algorithms 
can stem from various sources, including historical data, algorithmic design, and decision-making processes, thus necessitating the development of robust fairness metrics and frameworks to mitigate these disparities \cite{Ferrara_2023, bateni2022ai, ning2021shapley}.

%


Various metrics have been proposed to evaluate the fairness of credit scoring algorithms, encompassing disparate impact analysis, demographic parity, and equal opportunity criteria: disparate impact analysis examines whether the outcomes of the algorithm disproportionately impact protected groups; demographic parity ensures that decision outcomes are independent of demographic characteristics such as race, gender, or age; equal opportunity criteria focus on ensuring that individuals have an equal chance of being classified correctly by the algorithm, irrespective of their demographic attributes. Still, several challenges persist in implementing fair algorithms. One key challenge is the trade-off between fairness and predictive accuracy, as optimizing for one may inadvertently compromise the other. Moreover, biases inherent in training data, algorithmic design, and decision-making processes can perpetuate unfair outcomes, necessitating careful consideration and mitigation strategies.

The literature on fairness detection and mitigation in credit scoring has seen significant advancements, with researchers proposing various methods to address biases and promote equitable outcomes \cite{hardt2016equality,feldman2015certifying,zafar2017fairness,louizos2017variational,donini2020empirical, https://doi.org/10.1002/qre.3579}. Hardt et al. \cite{hardt2016equality} examine fairness in the FICO score dataset, considering race and creditworthiness as sensitive attributes. They employ statistical parity and equality of odds as fairness metrics to assess disparities in credit scoring outcomes across demographic groups. In \cite{feldman2015certifying}, Feldman et al. propose a fairness mitigation method based on dataset repair to reduce disparate impact, applying it to  the German credit dataset \cite{german_credit_ds}. They focus on age as the sensitive attribute and employ techniques to adjust the dataset to mitigate biases in credit scoring outcomes. Zafar et al. \cite{zafar2017fairness} introduce a regularization method for the loss function of credit scoring models to mitigate unfairness with respect to customer age in a bank deposit dataset. Their approach aims to prevent discriminatory outcomes by penalizing unfair predictions based on sensitive attributes. In \cite{louizos2017variational} the authors propose the implementation of a variational fair autoencoder to address unfairness in gender classification within the German dataset. Their approach leverages generative modeling techniques to learn fair representations of data and mitigate gender-based biases in credit scoring. In \cite{donini2020empirical}, Donini et al. analyze another regularization method aimed at minimizing differences in equal opportunity within the German credit ranking. Their empirical analysis highlights the effectiveness of regularization techniques in promoting fairness and equity in credit scoring outcomes. Most recently, the work in \cite{https://doi.org/10.1002/qre.3579} combines traditional group fairness metrics with Shapley values, though they admittedly may lead to false interpretations (cf. \cite{amoukou2022consistent}) and should thus combined with counterfactual approaches.

While the existing tools and studies present different fairness analyses and bias mitigation methods, to the best of our knowledge none of them enables the user to conduct an overall analysis yielding a combined and aggregated measure of the fairness violation risk related to all sensitive features selected. Moreover, our approach is model-agnostic --- while many others are not --- while still allowing for \emph{bias mitigation} considerations to be done.

We offer such a result using  \brio{}, a bias detection and risk assessment tool for ML and DL systems, presented in \cite{DBLP:conf/beware/CoragliaDGGPPQ23} and based on formal analyses introduced in \cite{DBLP:conf/atal/DAsaroP21,DBLP:conf/aiia/PrimieroD22,DBLP:journals/corr/abs-2302-00958,dasaro2024checking}.
In the present paper, we showcase its use on the UCI German Credit Dataset \cite{misc_statlog_(german_credit_data)_144} and present an encompassing, rigorous analysis of fairness issues within the context of credit scoring,
aligning with the recent ethical guidelines. To operationalize these principles, we measure the fairness metrics over the sensitive attributes present in the German Credit Dataset, 
quantifying and evaluating fairness across various demographic segments, thereby seeking to identify potential sources of bias and discrimination.

The rest of this paper is structured as follows. In Section \ref{sec:preliminary} we provide a preliminary illustration of the dataset under investigation, the features considered and the performance.
In Section \ref{sec:model} we explain how we constructed a ML model trained on the dataset considered for credit score prediction, its evaluation and validation and the results on score distribution. In Section \ref{sec:brio} we illustrate the theory behind bias identification and risk evaluation of \brio{}. In Section \ref{sec:risk} we present the results of risk evaluation on the UCI German Credit Dataset using \brio{}. We conclude in Section \ref{sec:conclusions} with further research lines.


\section{Preliminary Analysis}\label{sec:preliminary}\label{dataset}


The UCI German Credit Dataset stands as a cornerstone in the field of credit risk assessment research. This dataset offers a comprehensive compilation of attributes pertinent to creditworthiness evaluation, providing researchers with useful insights into the factors influencing lending decisions. The dataset comprises $1,000$ instances, each characterized by a set of $20$ input variables and an associated binary label representing the occurrence or not of the default event.

These input variables encompass demographic, financial, and credit-related attributes, including age, gender, marital status, duration of credit history, employment status, and housing situation, among others. Additionally, the dataset includes categorical and numerical features, facilitating diverse analytical approaches and modeling techniques. The binary label indicates whether or not a default has been observed in the credit history. This binary classification enables the evaluation of predictive models in terms of their ability to distinguish between creditworthy and non-creditworthy individuals, thereby serving as a benchmark for model performance assessment.

The UCI German Credit Dataset has garnered widespread attention within the research community due to its richness in features and relevance to real-world credit assessment scenarios. Researchers have utilized this dataset to develop and benchmark various machine learning algorithms for credit scoring, ranging from traditional logistic regression models to more sophisticated ensemble methods and neural networks.


Across the attributes provided by the German Credit Dataset, we have selected some and then formulated dependency among them, so as to represent sensitive classes for which fairness should be ensured. These sensitive attributes play a crucial role in assessing the impact of the credit scoring model on different demographic groups, thereby guiding efforts to mitigate potential biases and disparities. The sensitive classes we considered in our analysis are the following:
\begin{enumerate}
    \item Gender, categorized as male or female.
    \item Age Groups, segmented into age brackets including $[0\text{-}27]$, $[27\text{-}37]$, $[37\text{-}47]$, and $[>47]$.
    \item Foreign Flag, categorized as foreign worker or not foreign worker.
\end{enumerate}
These attributes represent diverse demographic characteristics that may influence creditworthiness assessment and are thus pivotal for ensuring fairness in lending practices.

\begin{figure}[t]
\centering
\begin{subfigure}{\textwidth}
\includegraphics[width=0.5\linewidth]{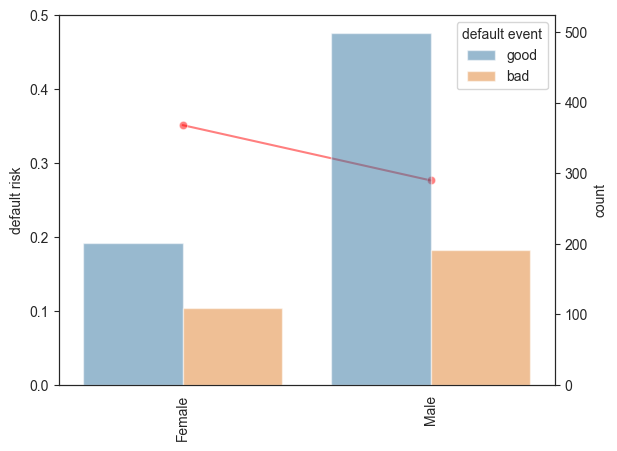}
\includegraphics[width=0.5\linewidth]{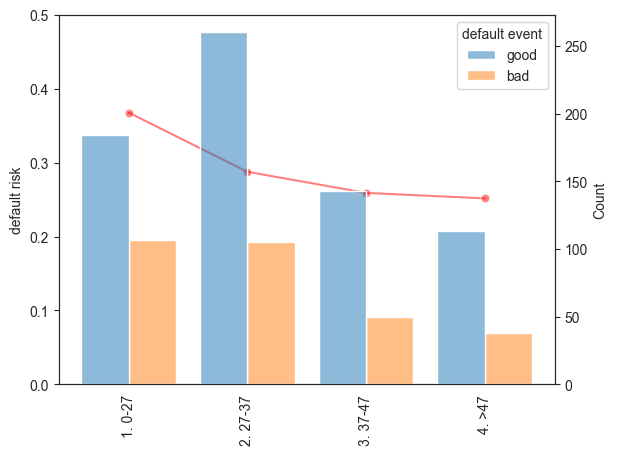}
\includegraphics[width=0.5\linewidth]{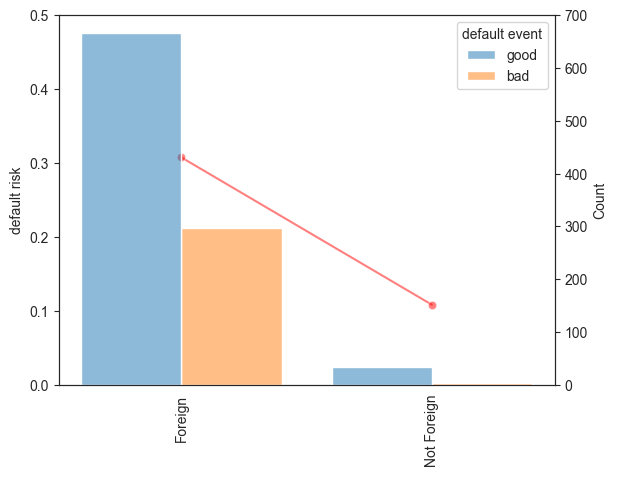}
\end{subfigure}
\caption{Default probability (red line, left vertical axis) and distributions (blue-orange bars, right vertical axis) for the sensitive variables.}
\label{performance_plot}
\end{figure}

To gain insights into the default distribution 
across sensitive classes within the input data, we compute the mean value of the default variable and represent it graphically in \Cref{performance_plot}. This visualization provides a comprehensive overview of the model's default probability across different demographic groups, facilitating the identification of potential disparities or biases.

From the default probability distribution, we observe notable variations across sensitive attributes. Specifically, we observe that:
\begin{enumerate}
    \item Males tend to exhibit better default risk outcomes compared to females.
    \item Older age groups demonstrate better default risk outcomes relative to younger age ones.
    \item Domestic workers exhibit better default outcomes compared to foreign workers.
\end{enumerate}

These findings underscore the importance of considering sensitive attributes in credit scoring models and highlight the necessity of addressing potential biases to ensure fairness and equity in lending decisions. We conduct an analysis to examine the impact of these sensitive attributes on the model's predictions and explore mitigation strategies to address any observed disparities. By incorporating fairness-aware techniques and algorithmic interventions, we aim to enhance the inclusivity and equity of the credit scoring model, ultimately fostering fair lending practices and promoting financial inclusion.

\section{ML model construction}\label{sec:model}

We construct an ML model for credit score prediction using the functions \texttt{BinningProcess} and \texttt{Scorecard} provided by the \texttt{Optibinning} Python library,  designed for optimal binning of continuous and categorical variables, tailored specifically for credit scoring and risk modeling applications, see \cite{navaspalencia2022optimal}. 

The first step in constructing our machine learning model involves the binning process, a crucial preprocessing step for transforming continuous variables into categorical bins. The \texttt{BinningProcess} function enables us to automatically identify and create optimal bins for each input variable based on statistical criteria such as entropy, $\chi^{2}$, or custom metrics. Using the UCI German Credit Dataset as input, the \texttt{BinningProcess} function partitions each continuous variable into a set of bins, optimizing the bin boundaries to maximize the predictive power of the resulting bins. By discretizing continuous variables into bins, we reduce the complexity of the input space and facilitate the interpretation of the model.

Once the binning process is complete, we proceed to generate a scorecard for credit scoring using the \texttt{Scorecard} function. A scorecard is a tabular representation of the credit scoring model, mapping each bin of the input variables to corresponding score points based on its predictive strength. The \texttt{Scorecard} function leverages the binned variables obtained from the binning process to compute the weight of evidence (WOE) and information value (IV) for each bin. These metrics quantify the predictive power and discriminatory ability of each bin in separating good and bad credit risks. Subsequently, the \texttt{Scorecard} function combines the WOE and IV values of all bins across the input variables to construct a unified scorecard, assigning score points to each bin based on its contribution to the predictive accuracy of the model. The resulting scorecard provides a transparent and interpretable framework for credit scoring, enabling lenders and analysts to assess the creditworthiness of applicants based on their respective scores.

Finally, we evaluate and validate the performance of the generated scorecard using appropriate metrics such as accuracy, area under the receiver operating characteristic (ROC) curve, and calibration measures. By assessing the model's discriminatory power, calibration, and stability over time, we ensure its reliability and robustness in real-world credit assessment scenarios.
\begin{figure}[t]
\centering
\begin{subfigure}{\textwidth}
\includegraphics[width=0.5\linewidth]{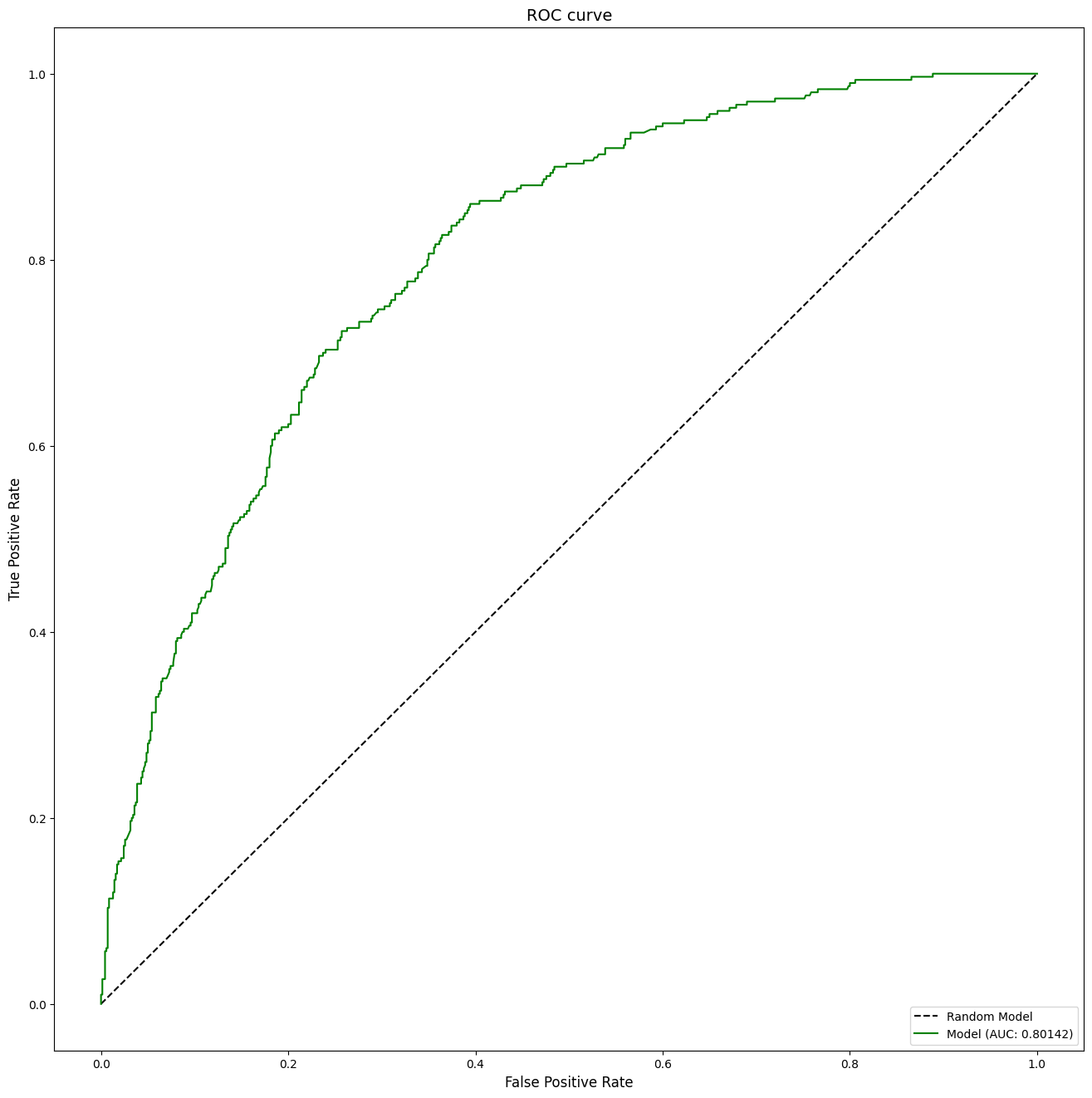}
\includegraphics[width=0.5\linewidth]{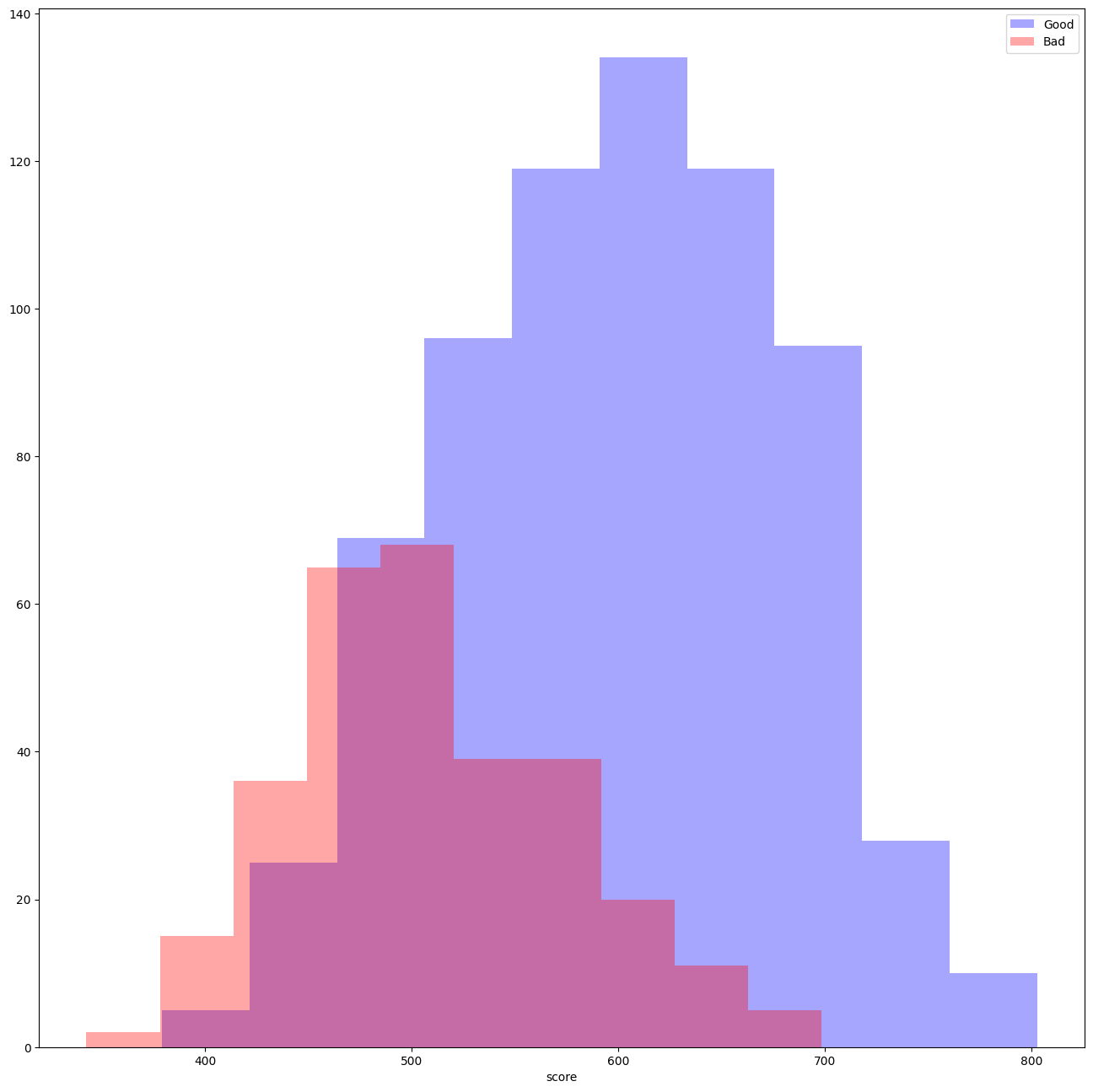}
\end{subfigure}
\caption{ROC curve of the model (left) and Good-Bad performance distributions relative to the predicted score.}
\label{roc_plot}
\end{figure}
In the realm of credit scoring, evaluating the discriminatory power of a model is paramount for assessing its effectiveness in distinguishing between creditworthy and non-creditworthy individuals. Common metrics used for this purpose include the Area Under the Curve (AUC) and the Gini index, derived from the Receiver Operating Characteristic (ROC) curve. The ROC curve is generated by plotting the True Positive Rate against the False Positive Rate at various threshold settings for classification. The AUC represents the area under this curve, quantifying the model's ability to rank individuals correctly. Additionally, the Gini index, calculated as the area between the ROC curve and the diagonal line (representing random chance), provides a measure of the discriminatory power of the model.

In our analysis, we computed the AUC and Gini index to assess the performance of the credit scoring model constructed using the \texttt{Optibinning} library. With an AUC of $0.8$ and a corresponding Gini index of $0.6$, our model demonstrates a good discriminatory capability. In the following analysis, we set a score threshold of $550$ to distinguish between ``Good'' ($\geq 550$) and ``Bad'' ($<550$), and with respect to that \emph{choice} we begin our fairness analysis.
Setting a threshold to decide who should be regarded as a good or bad payer is of course subject to discussion, and is in fact one of those actions that stakeholders can take to provide a fairer treatment with respect to some classes of individuals. We come back to said choice, and its relation to bias mitigation, in \Cref{sec:revenue}.

%
%

\begin{figure}[t]
\centering
\begin{subfigure}{\textwidth}
\includegraphics[width=0.5\linewidth]{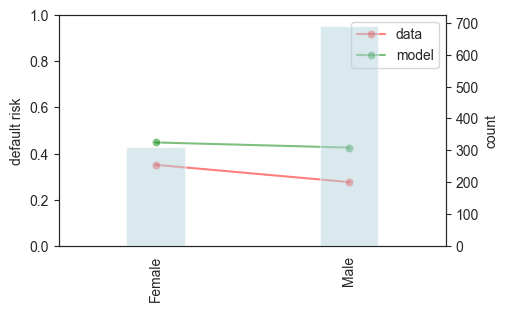}
\includegraphics[width=0.5\linewidth]{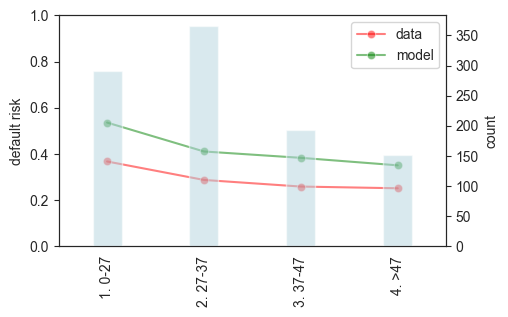}
\includegraphics[width=0.5\linewidth]{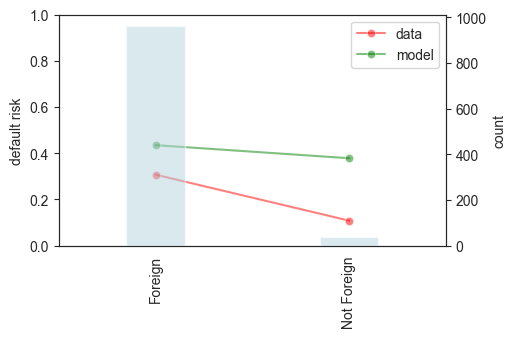}
\end{subfigure}
\caption{Comparison between the model's default risk (green line) and the data's default risk (red line) for the sensitive variables.}
\label{scorebin_plot}
\end{figure}
To visualize the distribution of predicted scores and compare the default probability between ``Good'' and ``Bad'' credit score categories, we can perform the same analysis as in \Cref{performance_plot}. The corresponding histograms are reported in \Cref{scorebin_plot}. This histogram provides insights into the model's ability to differentiate between creditworthy and non-creditworthy applicants based on their predicted scores, highlighting potential patterns or disparities in scoring outcomes. We note that the model's predictions essentially reflect what was found by the default analysis. However, the relative difference between the various sensitive classes, turns out to be attenuated in the case of foreign flag and accentuated in the case of gender and age groups.

\section{Fairness violation analysis in \brio{}}\label{sec:brio}

For the detection of fairness violations and consequent risk measurement we use \brio{}, a model-agnostic, bias and risk assessment tool designed to work on the I/O of ML and DL systems.\footnote{The open source code is available at \url{https://sites.unimi.it/brio/brio-x-alkemy/}. For a technical presentation of its features, and validation, we refer to  \cite{DBLP:conf/beware/CoragliaDGGPPQ23}.}
The module of \brio{} devoted to the detection of fairness violations takes as input the outputs of an AI model---encoded as a set of datapoints with relative features---and a set of parameters  including the designation of one or more \emph{sensitive features}, also called  \emph{protected attributes}. The output of the tool is an evaluation of the possibility that the AI model under consideration is unfair with respect to the designated features.
%
%
%

The system closely guides the user in the process of setting  parameters, 
and remains customisable with respect to the mathematical details of the analysis: the choices left to the user are those that actually make a conceptual difference in the outcome of the analysis, and implications of each choice are explained to the user along the way.

The system can conduct two kinds of analyses, consisting in comparing 
\begin{enumerate}
	\item the behaviour of the AI system against a desirable behaviour,
	
	\item the behaviour of the AI system with respect to a sensitive class $c_1\in F$ and another sensitive class $c_2\in F$ related to the same feature $F$.
\end{enumerate}

If the second analysis alerts of a possibly biased behaviour, it is possible to conduct a subsequent check on some (or all) of the subclasses of the considered sensitive classes. This second check is meant to verify whether the bias encountered at the level of the classes can be explained away by features of the individuals that are different from the sensitive feature at hand. 

Consider, for instance, the following situation. 

\begin{quotation}
Database $D$ contains details of individuals, with their age, gender, and level of education. Algorithm $A$ predicts the likelihood of default on credit and, for each datapoint, labels it as ``likely to default'' or ``not likely to default''. We wish to check if age is a sensitive factor in such prediction. We feed \brio{} with $D$ and the outputs of $A$ with respect to $D$. Suppose we consider the feature \texttt{age} as sensitive. \brio{} allows us to compare either how the behaviour of $A$ with respect to \texttt{age} differs from an ideal behaviour (in this case, we might consider ideal the case in which the frequency of elements which are labelled ``not likely to default'' is the same for each age group), or how different age groups perform with respect to one another.
\end{quotation}
 
The checks above are executed by comparing probability distributions indicating how probable it is that a generic element of a class is labelled in a certain way by the algorithm. In order to compute the difference between the behaviour of the AI system under investigation---described by the probability distribution $Q$---and another behaviour $P$ (either the ideal behaviour, when available, or the behaviour of the algorithm with respect to a different class) various means of comparison are employed by \brio, depending on the analysis one wishes to conduct. The divergence measures employed to compute the difference between two probability distributions are illustrated in the following.


\paragraph{Kullback-Leibler divergence.}
When we wish to compare how the system behaves with respect to an \textit{a priori} optimal behaviour $P$, we use the Kullback–Leibler divergence $D_{\mathrm{KL}}$:

\[D_{\mathrm{KL}}(P\parallel Q)= \sum _{x\in X} P(x)\cdot \log_2 \Big( \frac{P(x)}{Q(x)}\Big)\,.\]

This divergence was introduced in \cite{dkl} in the context of information theory, and it intuitively indicates the difference, in probabilistic terms, between the input-output behaviour of the AI system at hand and a reference probability distribution. It sums up all the differences computed for each possible output of the AI system weighted by the actual probability of correctly obtaining that output. Notice that this is not symmetric and takes values in $[0,+\infty]$: the asymmetry accounts for the fact that the behaviour we are monitoring is, in fact, non symmetric---as $P$ is a theoretical distribution that we know, or consider to be, optimal, while $Q$ is our observed one---while to make it fit the unit interval we adjust the divergence as follows.

\[D'_{\mathrm{KL}}(P\parallel Q)=1-\exp (- D_{\mathrm{KL}}(P\parallel Q)).\]

\paragraph{Jensen-Shannon divergence.}
When we wish to compare classes, instead, a certain symmetry is required. Hence, we employ  the Jensen-Shannon divergence 

\[D_{\mathrm{JS}}(P\parallel Q)=\frac{D_{\mathrm{KL}}(P\parallel M)+D_{\mathrm{KL}}(Q\parallel M)}{2},\]
with $M=(P+Q)/2$. This was introduced in \cite{lin_divergence} as a well-behaved symmetrization of Kullback-Leibler. It takes values in $[0,1]$.



%


When the comparison does not simply concern two classes since the values of the considered feature induce a more numerous partition, we need to aggregate the results obtained by computing the employed divergence on pairs of classes.


Suppose that we are studying the behaviour of the model with respect to the feature $F=\{c_1, \ldots , c_n\}$ which induces a partition of our domain into the classes $c_1, \ldots , c_n$. The first step is the pairwise calculation of the divergence with respect to the different classes induced by $F$. Hence, for each pair $(c_i, c_j)$ such that $c_i, c_j\in F$, we compute $D(c_i\parallel c_j) $ where $D$ is the  preselected divergence and consider the set $\{D(c_i\parallel c_j) : c_i,c_j\in F\ \&\ i\neq j\}$. For instance, if we are considering \texttt{age} as our feature  $F$ and we partition our domain into three age groups, we might have
\[F=\{\tt over\_50\_yo, between\_40\_and\_50\_yo, below\_40\_yo\}.\]
\brio{} enables us to choose between maximal, minimal or average divergence  to aggregate the obtained values.


In order to decide whether the behaviour of the AI system diverges  significantly on two classes -- or 
with respect to an ideal distribution -- a threshold is employed. If the divergence value is greater than the threshold, the output of \brio{} indicates that a fairness violation might have occurred. If, otherwise, the divergence value is smaller than the threshold, then the discrepancy in the behaviour of the AI system is deemed irrelevant and no violation is signalled. The value of the threshold can be changed depending on the case at hand, either by setting it manually or by leaving  \brio{} to select it automatically on the basis of the available data. In case the threshold is automatically computed by \brio{}, the threshold value $\varepsilon$ will be computed by a function with three parameters: $\varepsilon = f( r ,n_C , n_D)$. 

The parameter $r$ concerns the rigour required to analyse the case at hand and is selected by the user. Two settings are possible: 
\begin{itemize}
\item if $r=\mathtt{high}$, then the system will be extra attentive about the behaviour of the model in relation to it;
\item if $r=\mathtt{low}$, then the behaviour of the model with respect to the considered feature is considered significant only if it is particularly extreme.
\end{itemize}
This setting distinguishes between a thorough and rigorous investigation and a simple routine check.

The parameter $n_C$ is the number of classes related to the sensitive feature under consideration. We  call this the {\it granularity} of the classes related to the sensitive feature. When the classes under consideration are many, the divergences in the behaviour of the AI system can be of small entity but concern many classes. Hence, we need to be attentive also about small divergences and thus the threshold should be stricter.

Finally, the threshold is scaled with respect to the cardinality $n_D$ of the classes related to the sensitive feature under consideration. Large classes require a stricter threshold. This is a rather obvious choice related to the fact that statistical data related to a large number of individuals tend to be more precise and fine grained, as exemplified already above.


Formally, the threshold is computed as follows:
\[
\varepsilon = f(r,n_C, n_D) = (n_C \cdot n_D) \cdot m + (1-(n_C \cdot n_D)) \cdot M
\]
where $m$ is the lower limit of our interval (determined by the argument $r\in\{\mathtt{high}, \mathtt{low}\}$) and $M$ is its upper limit.

\section{Risk assessment in \brio}

The \brio{} system features a module devoted to the measurement of  risk associated with fairness violations by AI systems. The risk measure produced aggregates the results of all available relevant tests detecting fairness violations. The module takes in input a series of $n$ different test results, relative to possibly different sensitive features, and returns one value in the real unit interval $[0,1]$ which represents how high is the risk that the tested AI system behaves in an unfair manner.

As the \brio{} bias detection module does not only compare the behaviour of the AI algorithm on the classes relative to the sensitive feature, but can also execute similar checks on possibly several subclasses, the result of one fairness test will in general consist of $m$ lines. Each one of these lines will be relative to a subclass of the considered classes. Suppose, for example, that our sensitive feature is \texttt{gender}, the detection module of \brio{}  will not only compare the behaviour of the AI algorithm on the classes obtained by selecting a particular value of \texttt{gender}, but will also compare the behaviour of the AI algorithm on the subclasses obtained by fixing the values of features different from \texttt{gender} and by varying the value of \texttt{gender}. For instance, one line of the output will be about the behaviour of the AI system on the class of male individuals as compared to its behaviour on the class of female individuals, another line will be about the behaviour of the AI system on the subclass of rich male individuals as compared to its behaviour on the class of rich female individuals, yet another line will be about the behaviour on the subclass of poor male individuals as compared to its behaviour on the class of poor female individuals, and so on.


Therefore, each line of the output will provide the following information:
\begin{itemize}
\item the set of non-sensitive feature values used to determine the considered subclasses, if any;

\item the number of elements of the union of all considered (sub-)classes;

\item the value of the divergence for the considered (sub-)classes;

\item the threshold employed.
\end{itemize}

This information will be used to compute the overall risk measure emerging from a series of tests. In computing this measure, it is also possible to choose whether to focus on group fairness or individual fairness. Intuitively, focusing on {\it group fairness} means deeming more serious a discrimination based on very little information: for instance, a choice made only on the basis of the value of the sensitive feature will be a group discrimination. Focusing on {\it individual fairness}, on the other hand, means deeming more serious a discrimination between two individuals which have many values in common but a different value relatively to the sensitive feature. Abundant literature discussing their reciprocal incompatibility is available, see \textit{e.g.} \cite{10.1145/3351095.3372864, xu2024incompatibility}. \brio{} provides the option to choose either of the two.

Suppose now that $n$ tests are performed,\footnote{Either manually by the user, or according to an automatic \emph{strategy} depending on  need.} then the overall risk measurement function associated to a battery of tests can be formally defined as follows
\[\frac{1}{n}\cdot \sum ^n_{i=1}\mathrm{R}_i\]
with $\mathrm{R}_i$ the individual risk of each test computed.

Classically -- and informally -- the \emph{risk} associated to an event is considered to be proportional both to the \emph{likelihood} of its occurrence, and to the \emph{damage} that it might cause, i.e. it is given by the following product:
\[
\mathrm{R} = (\text{likelihood of failure at event}) \;\cdot\; (\text{damage of failure at event})\,
\]
With this intuition in mind, we define each $\mathrm{R}_i$, with $i\in\{1,\dots,n\}$, as
\[\mathrm{R}_i=\sum^m_{j=1} \delta (i,j)\cdot q(i,j)\cdot \sqrt[3]{\varepsilon (i,j)} \cdot\sqrt[3]{|e(i,j)|}\cdot w(i,j) \]where $m$ is the number of lines in the output of test $i$ and
\begin{enumerate}
\item $\delta (i,j)=1 $ if line $j$ of test $i$ is about a violation of fairness, and $\delta (i,j)=0 $ otherwise;

\item $q(i,j)$ is the number of elements in the union of the two classes (or subclasses) used for the comparison relative to line $j$ over the total number of datapoints;

\item $\varepsilon (i,j)$ is the threshold employed at line $j$;

\item $e(i,j)$ is the distance between the divergence and the threshold at  line $j$;

\item $w(i,j)$ is the weight of the possible fairness violation relative to line $j$.
\end{enumerate}
Intuitively, $\delta$, $q$ and $e$ account for the likelihood that a given line $j$ is flagged as a failure, and the weights $\varepsilon$ and $w$ determine the \emph{seriousness} of said failure.

In more detail, $\delta (i,j)$ simply sets the addend relative to line $j$ to $0$ if line $j$ does not correspond to a fairness violation, $q(i,j)$ makes the addend proportional to the number of individuals involved in the possible violation over all individuals, $e(i,j)$ makes the addend proportional to the gravity of the violation in terms of distance from the threshold, $\varepsilon (i,j)$ makes the addend inversely proportional to the strictness of the threshold employed. Factors involving $e$ and $\varepsilon$ are typically\footnote{Using the automated threshold they are, but it is always possible to use customized thresholds. Notice that the closer that is to $1$, the smaller is the effect of taking the cube root.} two or three orders of magnitude smaller than the others, so we scale their weight taking their cube root.

The weight $w(i,j)$  of the violation depends, in turn, on whether one focuses on {\it group fairness} or on {\it individual fairness}. In the first case, the weight increases if the possible fairness violation concerns a class determined by a few features (thus, a rather general class). In the second case, the weight increases if the possible fairness violation concerns a class determined by many features (thus, a rather specific class).

\section{Risk analysis via \brio{} for the German Credit Dataset}\label{sec:risk}

In this section we present the results of the application of \brio{} to the analysis of risk for the German Credit dataset. As explained above, \brio{}'s module for risk analysis works by aggregating several results of the fairness detection tool. The values obtained by several, individual checks on fairness -- possibly performed on different sensitive features -- are combined into a unique value indicating the global risk related to the fairness of the AI model. 

We first need to select some sensitive features. We selected three: {\it gender}, {\it nationality} and {\it age}. Moreover, the detection tool can conduct a series of double-checks on subclasses of the classes determined by the sensitive features. To this aim, some non-sensitive features are selected to determine the considered subclasses. 
%
%
%
%
The choice of these non-sensitive features is guided by the relevance that they bear with respect to the output of the AI model and by the legitimacy of their usage as criteria to be used in the prediction. The particular selection that we made here is the following: \texttt{`Attribute1'} (status of existing checking account), \texttt{`Attribute3'} (credit history), \texttt{`Attribute6'} (existence of savings account or bonds), \texttt{`Attribute10'} (existence of debtors or guarantors), \texttt{`Attribute12'} (properties owned), \texttt{`Attribute14'} (existence of other instalment plans). 
These features are all connected to the financial history of the subject input of prediction and their values constitute reasonably legitimate criteria for predicting a credit risk category for the considered subject.

 \
\begin{table}[t]
\centering
\begin{center}
\begin{tabular}{ |p{2cm}||p{3cm}|p{3cm}| p{2cm}|}
 \hline
 Sensitive  & Hazard value     & Hazard value          & Risk value \\
 feature    & (group fairness) & (individual fairn.) & 
 \\\hline\hline
 Gender       & $0.00226$ & $0.00232$  & 
 \multirow{3}{4em}{$0.00584$}\\\cline{1-3}
 Age      &  $0.00946 $& $0.00946$&\\\cline{1-3}
 Nationality & $0.00720$& $0.00720$&\\
 \hline

\end{tabular}
\end{center}
\caption{Hazard values for group fairness and BRIO risk. For these measures, we selected the Jensen-Shannon divergence as the distance function and set the threshold to the "high" level.}
\label{fairness_table}
\end{table}
\

Some examples of the partial results---technically called {\it hazard} values---used to compute the final outcome of the risk analysis  and the final risk value obtained by aggregating all these hazard values are displayed in Table \ref{fairness_table}. Notice that, in computing the hazard values, it is possible to employ different selections of non sensitive features to conduct double checks on the fairness violation detection. In this case, we employed the same list of  non sensitive features for the double checks since they all seem relevant with respect to all sensitive features investigated. In order to compute these values, we employed the automatic threshold calculator of the \brio{} tool -- presented at the end of Section \ref{sec:brio} -- and set the sensitivity of the threshold to {\texttt high} (low tolerance to fairness violations). As shown in Table \ref{fairness_table}, these values constitute aggregations of several bias detection tests concerning both {\it group fairness} and {\it individual fairness}.


Let us briefly discuss the obtained values. First, there is a considerable difference between the hazard values obtained for tests on gender and those for the tests on nationality and age. Cases like this clearly call for further, localized analyses. 
Specific tests can be conducted by the different modules of \brio{} in order to explain in even more detail the problem encountered in the global risk evaluation.
%
%
For instance, some runtime warnings returned in the output of the hazard computation 
for group fairness on nationality
have signaled an uneven distribution of the elements of the data frame with respect to the different possible values of the nationality feature: some classes related to this feature are empty. Hence, the differences in the hazard values related to gender and nationality can be motivated by the fact that, while the undesired behaviour of the AI model with respect to gender can be partially explained out if we consider the distribution of gender classes with respect to the subclasses induced by the combination with the considered non sensitive features, the undesired behaviour of the AI model with respect to the nationality of the subject cannot be  explained out in the same way. And this is in turn due to the fact that the instances in the database belonging to different nationality classes are not evenly distributed among the classes induced by the non sensitive features.


The final risk value obtained  does not seem to indicate extreme unfairness. It looks, nonetheless, non negligible. Obviously, this value only assumes its full meaning only in comparison to those obtained by similar analyses for other databases, models, or classification threshold choices. The latter case is precisely the one we are considering next
%
when applying the risk measures 
to the default event of the German Credit dataset. While keeping gender, age, and nationality as sensitive classes, we employ the risk metric not in relation to the model output but rather to the dataset's default attribute. This approach yields an estimation of the inherent level of unfairness present in the input data.
If the various categories within a sensitive class were statistically equally represented, such unfairness might be deemed somewhat acceptable, as it could stem from genuine actual disparities in the credit behavior among these categories. However, if there are significant imbalances in representativeness among the various categories, the reliability of the risk measure conducted in relation to default diminishes. It is therefore crucial to ensure that the model does not introduce a higher risk than what is inherently present in the data, meaning that the model should not be more discriminatory than its input. 

 \
\begin{table}[t]
\centering
\begin{center}
\begin{tabular}{ |p{3cm}||p{3cm}|p{3cm}| }
 \hline
 Sensitive class& Data hazard value &  Model hazard value \\
 \hline\hline
 Gender   &  $0.00165 $    & $0.00226$\\\hline
 Age     &$0.00617 $ & $0.00946 $\\\hline
 Nationality&   $0.01054$  & $0.00720$\\
 \hline
\end{tabular}
\end{center}
\caption{Comparison between data and model hazard values for group fairness.}
\label{parity_minority_table}
\end{table}
\

The BRIO risk computation with respect to default turns out to be $0.00566$. In Table \ref{parity_minority_table}, we compare the hazard values obtained for the model output and the input performance attribute. When comparing these results, it becomes apparent that there is a global $0.02 \%$ risk difference between the model and data. This indicates that the model is generally fair and does not introduce significant additional bias beyond what is present in the data. However, examining individual sensitive classes reveals that the difference in hazard values is more pronounced for age. In contrast, for the nationality, the difference is negative, suggesting that the model effectively corrects the minor bias present in the data.

\section{Revenue analysis}\label{sec:revenue}

In addition to the evaluation of the discriminatory power and predictive accuracy of the credit scoring model, a possible further application of the presented methods is the analysis of the interplay between revenue generation and fairness risk management. We can conduct a quantitative study of the effects on revenue generation that the choices related to the management of this kind of risk can have. In order to do this, we limit our focus to data with good predicted scores only. Two key metrics used for our purpose are \emph{provisions} and \emph{bad rate}, which provide insights into the financial implications of lending decisions.


Provision refer to the funds set aside by financial institutions to cover potential losses arising from non-performing loans or defaults. By accurately predicting credit risk and identifying high-risk applicants, the credit scoring model enables lenders to allocate provisions more effectively, mitigating the impact of defaults on their balance sheets. In our analysis, we evaluate the provisions allocated based on the predicted credit scores generated by the model.

The bad rate (BR), also known as the default rate, measures the proportion of loans that become non-performing or default within a specified period. A lower bad rate indicates a lower incidence of defaults, reflecting the effectiveness of the credit scoring model in identifying creditworthy applicants and mitigating credit risk. By analyzing the bad rate across different risk segments defined by the credit scoring model, we can assess its ability to differentiate between high-risk and low-risk applicants. A model that accurately predicts credit risk should exhibit a higher bad rate among high-risk applicants and a lower bad rate among low-risk applicants, enabling lenders to make informed lending decisions and minimize default risk.

The provision and the bad rate are not independent quantities and they can be related upon assumptions or simplifying hypotheses. If we consider the total credit amount ($TCA$), by assuming a fixed fraction of defaults for customers with poor default outcome, we can estimate provisions using the following simple expression:
\begin{equation}
   \texttt{ provisions} = TCA \cdot BR \cdot 0.2
\end{equation}
where the factor 0.2 serves as an estimate for the fraction of expected defaults among customers with a default history.  This simplification enables us to quantify provisions based on the observed bad rates, providing a straightforward metric for risk assessment. It is important to acknowledge that reducing the expected default to a single fixed fraction entails strong assumptions, aimed at simplifying the analysis for practical purposes. In a real-world context, numerous factors must be considered to accurately estimate provisions and assess credit risk. These factors may include:
\begin{enumerate}
    \item segmentation of financial products, as different products may exhibit varying default rates, necessitating tailored provisions calculations for each segment;
    \item interest rates, as the cost of credit and associated interest rates can influence default probabilities and provisioning requirements;
    \item type of company involved, as corporate borrowers may present different risk profiles compared to individual consumers, impacting default likelihood and provisioning strategies;
    \item credit duration, as longer credit durations may entail higher default risks, necessitating adjustments to provision estimates;
    \item institution's credit policies, as lending institutions may have varying risk appetites and credit assessment methodologies, influencing provisioning practices.
\end{enumerate}
The profit derived from extending credit to customers with good scores is evidently determined by the sum of credits multiplied by their respective interest rates (IR). Therefore, the final profit is obtained as the difference between this revenue and the provisions.
\begin{equation}
    \texttt{profit} = \sum_{i} TCA_i \cdot IR_i - \texttt{provisions}
\end{equation}
where the sum is extended to the accepted customers (good score) without observed default.

\begin{figure}[t]
\centering
\includegraphics[width=1.1\linewidth]{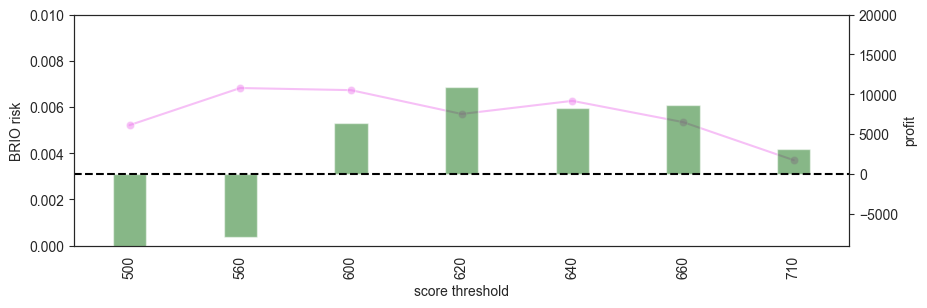}
\caption{Trends of profit (green bars, right vertical axis) and the model-data risk difference (pink line, left vertical axis) for multiple score thresholds.}
\label{revenue_plot}
\end{figure}

Both fairness risk and profit are to some extent dependent on the score threshold set for distinguishing between good and bad scores. Figure \ref{revenue_plot} illustrates their trends for various threshold values. Notably, both fairness and profit exhibit non-monotonic behaviour. This suggests that one approach to mitigating fairness risk could be to 
change the acceptance threshold in order to lower the risk measure while still maintaining an acceptable level of profit. To this aim, a risk analysis as the one performed by the \brio{} tool in the previous section can be crucial to identify the best balance between fairness and profit. In particular, in Figure \ref{revenue_plot} we can see that setting the threshold around $620$ can, in this particular case, strike a very good balance between risk of fairness violations and profit. Further investigations by the different modules of \brio{} can also be used to understand more in depth what are the classes that are unfairly excluded from credit.


\section{Conclusions}\label{sec:conclusions}

We have presented a study in the context of credit scoring relying on the use of the \brio{} tool for the detection of fairness violations and for the measurement of the risk associated to them. These methods have been displayed as means to guide the alignment with recent guidelines on AI in the credit domain. As a case study, we have focused on the German Credit dataset, for which we have developed a machine learning model to predict credit risk scores. Among the variables in the dataset, gender, age, and nationality were identified as sensitive classes.

The \brio{} tool allows to compare the model's treatment of these sensitive classes. Additionally, we have introduced an associated new metric for measuring overall risk, which provides a comprehensive assessment by considering all potential sources of fairness violations generated by \brio. This metric offers a unique measure for evaluating the model with respect to bias amplification. Results obtained from applying these metrics indicate that the model built using the German Credit data is sufficiently fair, as it does not introduce significant bias beyond what is observed within the data. 

Finally, we have showcased a further possible application of the presented methods concerning the analysis of the effects that fairness risk management can have on revenue generation.

These findings underscore the importance of incorporating fairness considerations into credit scoring models and highlight the potential of innovative metrics to provide an integrated evaluation of model fairness. Future work will extend these approaches to other datasets and contexts, further refining the tools and methods used to ensure fairness in AI-driven credit scoring systems.

\bibliographystyle{alpha}
\bibliography{bib-german-credit}

\end{document}